\documentclass[runningheads]{llncs}

\newif\ifarxiv
\IfFileExists{./arxiv.txt}{\arxivtrue}{}

\usepackage{graphicx}
\usepackage{comment}
\usepackage{amsmath,amssymb}
\usepackage{color}

\usepackage[usenames,dvipsnames,table]{xcolor}
\usepackage{amsfonts}
\usepackage{float}
\usepackage{adjustbox}
\usepackage{multirow}
\usepackage{booktabs}
\usepackage[pagebackref=true,breaklinks=true,letterpaper=true,colorlinks,bookmarks=false]{hyperref}

\usepackage[inline, shortlabels]{enumitem}
\setlist[itemize]{leftmargin=*, noitemsep, topsep=0pt}
\setlist[enumerate]{label=\roman*), noitemsep, topsep=0pt}
\usepackage{array}
\usepackage{caption}

\newcommand{\mypar}[1]{\noindent{\bf #1.}}
\newcommand{\myparnodot}[1]{\noindent{\bf #1}}

\newcommand{\supervised}[1]{{\color{NavyBlue}{#1}}}
\newcommand{\proxy}[1]{{\color{RedViolet}{\bf #1}}}
\newcommand{\target}[1]{{\color{OliveGreen}{\bf #1}}}

\newcommand{\kmeans}{$k$-means}

\newcommand{\bert}{BERT$_\text{base}$}
\newcommand{\tfm}{\texttt{tfm}}
\newcommand{\att}{\texttt{att}}
\newcommand{\fc}{\texttt{fc}}
\newcommand{\tp}{\texttt{tp}}
\newcommand{\tpstar}{{TP}$_\star$}
\newcommand{\tplabel}{TP\textsubscript{Label}}
\newcommand{\tppostag}{{TP\textsubscript{Postag}}}
\newcommand{\tpcluster}{{TP\textsubscript{Cluster}}}
\newcommand{\tplda}{{TP\textsubscript{LDA}}}
\newcommand{\icmlm}{{ICMLM}} 
\newcommand{\icmlmstar}{{ICMLM}$_\star$} 
\newcommand{\icmlmtfm}{{ICMLM}$_\text{\tfm}$}
\newcommand{\icmlmattfc}{{ICMLM}$_\text{\att-\fc}$}

\newcommand{\mphi}{\phi}
\newcommand{\mpsi}{\psi}
\newcommand{\matt}{p_\text{\att{}}}
\newcommand{\mtp}{\text{\tp{}}}
\newcommand{\mnorm}{\text{norm}}
\newcommand{\mlossmlm}{\ell_\text{mlm}}
\newcommand{\mlosstp}{\ell_\text{tp}}
\newcommand{\mlossicmlm}{\ell_\text{icmlm}}
\newcommand{\dx}{d_\text{x}} 
\newcommand{\dw}{d_\text{w}} 
\newcommand{\dz}{d_\text{z}} 
\newcommand{\hx}{H}
\newcommand{\wx}{W}
\newcommand{\nk}{K} 
\newcommand{\D}{\mathcal{D}} 
\newcommand{\C}{O} 
\newcommand{\X}{\vec{X}} 
\newcommand{\W}{\vec{W}} 
\newcommand{\V}{\vec{V}} 
\newcommand{\cent}{\xi} 

\usepackage{xspace}
\makeatletter
\DeclareRobustCommand\onedot{\futurelet\@let@token\@onedot}
\def\@onedot{\ifx\@let@token.\else.\null\fi\xspace}

\def\eg{\emph{e.g}\onedot} 
\def\ie{\emph{i.e}\onedot} 
 \def\vs{\emph{vs}\onedot}

\def\eq#1{Eq.~(\ref{#1})}
\def\fig#1{Fig.~\ref{#1}}
\def\tab#1{Tab.~\ref{#1}}
\def\sect#1{Sec.~\ref{#1}}

\newcommand{\Real}[1]{\mathbb{R}^{#1}}
\renewcommand{\vec}[1]{\mathbf{#1}}
\DeclareMathOperator*{\argmin}{arg\,min}
\DeclareMathOperator*{\argmax}{arg\,max}
\DeclareMathOperator*{\E}{\mathbb{E}}

\newcommand*{\@rowstyle}{}
\newcommand*{\rowstyle}[1]{
  \gdef\@rowstyle{#1}%
  \@rowstyle\ignorespaces%
}
\newcolumntype{=}{
  >{\gdef\@rowstyle{}}%
}
\newcolumntype{+}{
  >{\@rowstyle}%
}

\makeatother

\begin{document}
\pagestyle{headings}
\mainmatter
\def\ECCVSubNumber{498}

\title{Learning Visual Representations \\ with Caption Annotations}
\titlerunning{Learning Visual Representations with Caption Annotations}

\author{Mert Bulent Sariyildiz, Julien Perez, Diane Larlus}
\index{Sariyildiz, Mert Bülent}
\authorrunning{M. B. Sariyildiz, J. Perez, D. Larlus}
\institute{NAVER LABS Europe}
\maketitle

\begin{figure*}[h!]
    \centering
    \includegraphics[width=\linewidth]{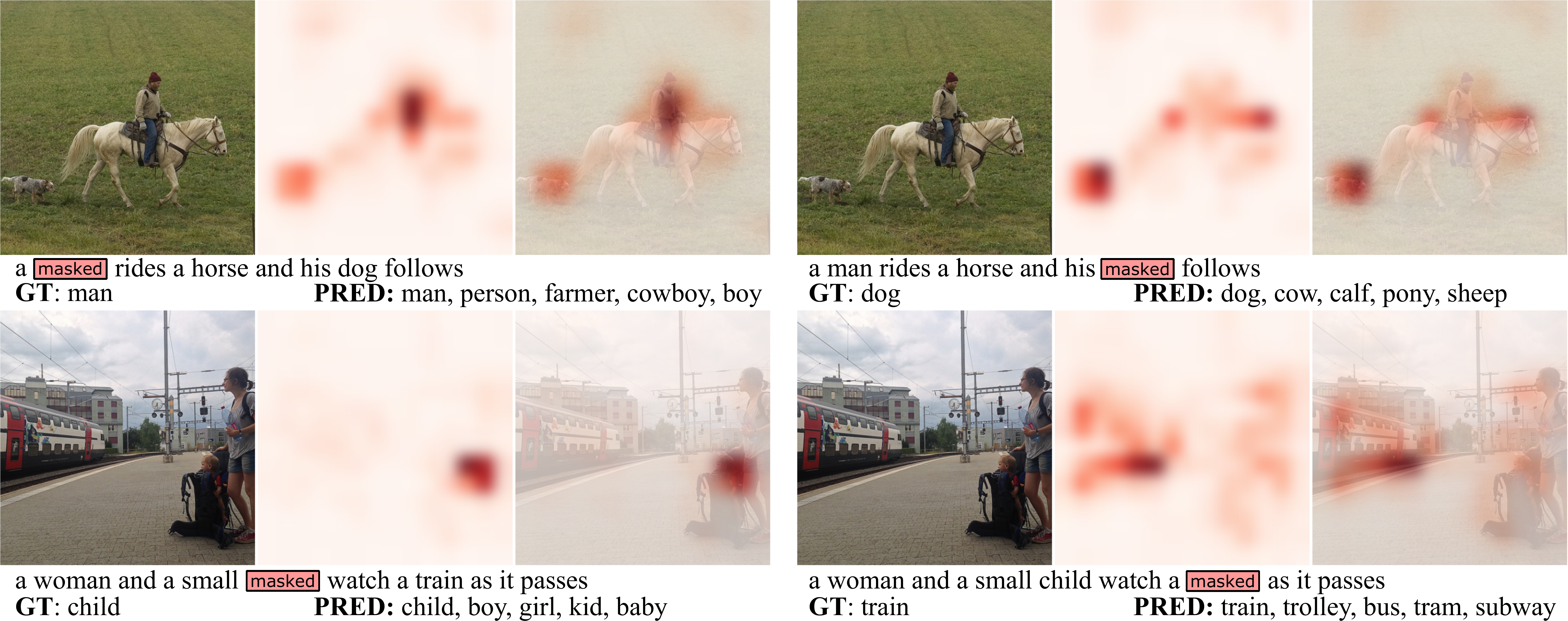}
    \caption{
    We introduce \textit{image-conditioned masked language modeling} (ICMLM), a proxy task to learn visual representations from scratch given image-caption pairs.
    This task masks tokens in captions and predicts them by fusing visual and textual cues.
    The figure shows how the {\em visual attention} changes as we mask different tokens in a caption (produced by our \icmlmtfm{} trained on COCO).
    }%
    \label{fig:front}
\end{figure*}

\begin{abstract}
Pretraining general-purpose visual features has become a crucial part of tackling many computer vision tasks.
While one can learn such features on the extensively-annotated ImageNet dataset, recent approaches have looked at ways to allow for noisy, fewer, or even no annotations to perform such pretraining.
Starting from the observation that captioned images are easily crawlable, we argue that this overlooked source of information can be exploited to supervise the training of visual representations.
To do so, motivated by the recent progresses in language models, we introduce {\em image-conditioned masked language modeling} (ICMLM) -- a proxy task to learn visual representations over image-caption pairs.
ICMLM consists in predicting masked words in captions by relying on visual cues. 
To tackle this task, we propose hybrid models, with dedicated visual and textual encoders, and we show that the visual representations learned as a by-product of solving this task transfer well to a variety of target tasks.
Our experiments confirm that image captions can be leveraged to inject global and localized semantic information into visual representations.
Project website: \url{https://europe.naverlabs.com/icmlm}.
\end{abstract}

\section{Introduction}%
\label{sec:intro}%
Large-scale manually annotated datasets~\cite{deng2009imagenet,zhou2017places} have been fueling the rapid development of deep learning-based methods in computer vision.
Training supervised models over such datasets not only leads to state-of-the-art results, but also enables networks to learn useful image representations that can be exploited on downstream tasks.
However, this approach has major limitations.
First, the cost and complexity of annotating datasets is considerable, especially when the class taxonomy is fine-grained requiring expert knowledge~\cite{deng2009imagenet,parkhi12cats,wah2011CUB}.
Second, retraining from scratch dedicated models for every new task is inefficient.

Some alternative approaches address these issues and require less curated or fewer annotations~\cite{mahajan2018ECCV,yalniz2019billion}.
At the other extreme of visual representation learning, self-supervised learning methods~\cite{caron19unsupervised,doersch2015context,doersch2017mtself,fernando17self,zhang2016colorful,zhang2017splitbrain}
do not require annotations and fabricate proxy labels from the data itself.
They induce regularities of the data itself, decorrelated from any specific downstream task annotations.
Unfortunately, recent findings show that these approaches are not data efficient, \ie~they require either extremely large training sets (up to a hundred million images)~\cite{caron19unsupervised,goyal19scaling} or need to be trained much longer with larger networks to express their full potential~\cite{chen20simclr,he20moco}.
Hence they demand huge computational resources.

Interestingly, data often comes with informative metadata for free.
For instance, user tags associated with images can be used as image labels~\cite{joulin2016learning,mahajan2018ECCV}.
Even richer, companion text for images, is sometimes available for free.
Using recent sanitation procedures~\cite{qi2020imagebert}, high-quality large-scale captioned datasets can automatically be constructed.

In this paper, we argue that learning visual representations with captions should significantly reduce the scale of the training sets required for pretraining visual representations.
If no text is available, in some context it is still easier to acquire short captions than expert-quality-level fine-grained class labels over thousands of categories like in ImageNet~\cite{deng2009imagenet}.
Yet, caption annotations have rarely been used to train visual representations from scratch.
Notable exceptions are~\cite{gomez17self,joulin2016learning,quattoni07learning} which learn image features by training to predict words in a caption or topic probabilities estimated from an associated text.
However, none of these approaches use the structure of the entire sentences, \ie~they treat words individually.
Recent studies~\cite{devlin2019bert,peters18deep} have shown the superiority of word representations which are conditioned by their surrounding, where the same word has different representations depending on the sentence.
We believe such caption representations should also be beneficial for learning image representations.

This paper focuses on the following research questions.
{\em Can we train transferable visual representations from limited sets of image-caption pairs?}
If so, {\em how should we formulate the interaction between images and captions?}
To address these questions, we propose several proxy tasks involving images and their captions which we use to train visual representations from scratch.
The first one (Sec.~\ref{sec:method_tp}) is intuitive and requires only extracting {\em image tags} from captions.
We propose several ways to do so, and we show that predicting image tags is already competitive compared to other pretraining strategies.
Then, to utilize the captions more effectively, and inspired by the recent advances in natural language processing~\cite{devlin2019bert}, we propose a second proxy task (Sec.~\ref{sec:method_icmlm}) which employs masked language modeling to learn visual representations.
Similar to the first proxy task, it also leverages both images and captions, but it additionally allows visual representations to learn to {\em localize semantic concepts} in captions.
Qualitative results show that the architecture proposed to tackle this second proxy task effectively leverages the text and attends to relevant image regions (see~\fig{fig:front}).

Our contributions are threefold.
First, we empirically validate that simple tag prediction tasks, where tags are obtained from captions, already learn transferable visual representations.
Second, in an attempt to benefit from captions more, we introduce a new task called {\bf i}mage-{\bf c}onditioned {\bf m}asked {\bf l}anguage {\bf m}odeling (ICMLM) and propose two multi-modal architectures to solve this task.
Third, we show that solving ICMLM leads to useful visual representations as a by-product.
These visual representations, which we obtain using only a hundred thousand captioned images, are competitive with recent self-supervised approaches leveraging a hundred million images, and, in some cases, even fully-supervised approaches showing how powerful a cue text is.

\section{Related Work}%
\label{sec:relwork}%
Pretraining CNNs on an external dataset has become standard practice in computer vision~\cite{chen18deeplab,gordo2017end,ren2015faster,sariyildiz2019CVPR}, especially for domains or tasks for which data is scarce.
The most common strategy is to train a CNN for the ImageNet-1K classification task~\cite{ilsvrc15} and then to use it as a feature extractor or to fine-tune it on a target task or domain.
Although this scheme has proven to be quite useful, designing fully-annotated datasets represents a significant effort requiring prior knowledge and domain expertise~\cite{deng2009imagenet}.
Thus, alternative research directions have gained interest.
We review the ones closest to our work.

\mypar{Weakly/Webly-supervised learning}
Two main research lines have prospered recently.
The first line focuses on using {\em metadata} associated to web data, such as tags or captions for images or videos~\cite{thomee15yfcc100m}.
Although the signal-to-noise ratio of samples crawled from the web may arguably be lower than carefully-constructed datasets, significant progress has been made leveraging this type of data to pretrain models~\cite{chen2015webly,hong2017wsseg}.
Among those, to learn visual representations,~\cite{joulin2016learning} extracts
the most common hashtags and words from the captions and titles of 99 million
images in the YFCC100M dataset~\cite{thomee15yfcc100m} and train to predict these words using CNNs.
Similarly,~\cite{mahajan2018ECCV} uses hashtags associated with images from Instagram to construct datasets containing up to 3.5 billion images.

The second line upscales ImageNet. Leveraging ImageNet labels, those approaches
produce {\em pseudo-labels} for additional unlabeled images~\cite{xie20self,yalniz2019billion}.
We note that these methods require initial annotations and extremely large-scale sets of images.
In contrary, our models need far less images, 118 thousand images at most, but companion captions to learn visual representations.

\mypar{Unsupervised representation learning}
Self-supervised approaches build a {\em pretext task} to learn image representations which are decorrelated from any downstream task and they do not require any manual annotations.
Often, {\em proxy tasks} consist in predicting missing pieces in manipulated
images, for instance context prediction~\cite{doersch2015context},
colorization~\cite{deshpande15learning,larsson17colorization,zhang2016colorful},
inpainting of missing portions~\cite{pathak2016contextenc}, prediction of image rotations~\cite{gidaris2018ICLR}, spotting artifacts~\cite{jenni18self}, or cross-channel prediction~\cite{zhang2017splitbrain}.
Besides, recently, contrastive learning-based unsupervised methods~\cite{bachman19learning,he20moco,oord2018cpc,wu18unsupervised} have showed significant improvements.
However, computational and data efficiency of these methods are still inferior to supervised models.

It is important to note that most unsupervised approaches are trained on {\em curated datasets} such as ImageNet for which images were carefully selected to form a well-balanced collection for a diverse set of fine-grained categories.
Although these approaches do not directly use ImageNet labels, they implicitly benefit from this careful selection and the resulting underlying structure of the dataset.
Indeed,~\cite{caron2018ECCV,doersch2015context} show that the feature quality drops when raw data are used instead of ImageNet.
Yet, assuming that a curated dataset such as ImageNet is readily available is a strong assumption.
Consequently, some works~\cite{caron19unsupervised,goyal19scaling,mahendran18cross} have evaluated unsupervised methods trained on {\em uncurated data}~\cite{thomee15yfcc100m}.
They have concluded that large amounts of raw data (\eg~96 millions images) is required to express the full potential of these approaches.
In this work, we focus on learning from a much smaller set of images by leveraging textual information.

\mypar{Vision and language}
Vision and language (VL) have been jointly leveraged to learn cross-modal representations for various VL tasks, such as cross-modal retrieval~\cite{gomez19self,wang16learning}, visual question answering~\cite{goyal17making}, captioning~\cite{sun2019vidbert} or visual grounding~\cite{deng18visual,hu2016natural}.
Building on the recent advances in natural language processing~\cite{devlin2019bert,Vaswani2017AttentionIA}, several works have fine-tuned BERT~\cite{devlin2019bert} to fuse visual and textual information~\cite{lu2019vilbert,su2020vlbert,sun2019vidbert,tan2019lxmert,zhou20unified} for VL tasks.
However, while learning cross-modal representations, such approaches rely on pretrained feature extractors, \ie~they use visual features pooled from regions of interest produced by a state-of-the-art detector such as Faster-RCNN~\cite{ren2015faster}.
Therefore, their objectives are formulated under the assumption that discriminative visual features are readily available for a list of relevant objects.
We note that such feature extractors are already state-of-the-art for most vision tasks, requiring {\em expensive bounding box annotations} to train.
Our approach follows a different path.
We focus on learning visual representations {\em from scratch} for purely visual tasks by leveraging captions.

\mypar{Learning visual features using text}
Only few works have taken advantage of the companion text to learn image representations.
\cite{quattoni07learning} creates and solves auxiliary prediction tasks from images with associated captions.
\cite{li2017ngrams} constructs label sets out of caption $n$-grams, and trains
CNNs by predicting these labels.
\cite{gomez17self} extracts topic models for Wikipedia pages using latent Dirichlet allocation and trains a CNN to embed their associated images in this topic space.
\cite{gordo17beyond} uses captions to learn image representations for the specific task of semantic retrieval.

We argue that language has a complex structure which cannot be reduced to computing n-grams statistics in a text.
Motivated by this, we differ and propose to use a pretrained language model -
which can be trained in an unsupervised manner for large text corpora - to represent captions and individual words in them.
In our experiments, we show that by doing so it is possible to learn visual representations that are useful for a broad range of tasks.

\section{Method}%
\label{sec:method}%
We argue that captions associated with images can provide semantic information about some {\em observable concepts} that can be captured by image representations.
Such concepts can be objects, attributes, or actions that visually appear in images.
With this motivation, given a dataset composed of image-caption pairs, we want to formulate non-trivial proxy tasks conditioned on both images and captions such that solving these tasks produce generic visual representations as a by-product.
In particular, we want such tasks to properly use the structure of caption sentences, and not only treat them as orderless sets of words.

To this end, we propose two proxy tasks focusing on two distinct objectives to train CNNs to recognize a predefined set of concepts in images.
The first proxy task captures {\em global} semantics in images by predicting image-level tags and is presented in \sect{sec:method_tp}.
The second proxy task, the image-conditioned masked language modeling task, focuses on {\em local} semantics in images and is detailed in \sect{sec:method_icmlm}.
Experiments show that both proxy tasks are complementary.

\mypar{Notations}
We assume that our dataset $\mathcal{D} = \{ (I^i, c^i)\}_1^N$ is composed of $N$ image-caption pairs.
We denote by $\C = \{o_i\}_1^K$ the set of concepts to be recognized in images.
As there can be multiple concepts in an image, we use binary label vectors $\vec{y} \in \{0, 1\}^{K}$ to denote the presence of concepts in images, \ie~$\vec{y}_k = 1$ if concept $o_k$ appears in image $I$ and $0$ otherwise.
We define two parametric functions $\mphi$ and $\mpsi$ which respectively embed images and text.
More precisely, $\mphi: I \rightarrow \X \in \Real{\hx \times \wx \times \dx}$ takes an image $I$ as input and produces $\X$  which is composed of $\dx$-dimensional visual features over a spatial grid of size $\hx \cdot \wx$.
Similarly, $\mpsi: c \rightarrow \W \in \Real{T \times \dw}$ transforms a caption (a sequence of $T$ tokens) into a set of $\dw$-dimensional vectors, one for each token.
In our models, we train only $\mphi$, which is a CNN producing visual representations, and we use a pretrained language model as $\mpsi$ that we freeze during training.

\begin{figure}[t!]
	\centering
	\includegraphics[width=\linewidth]{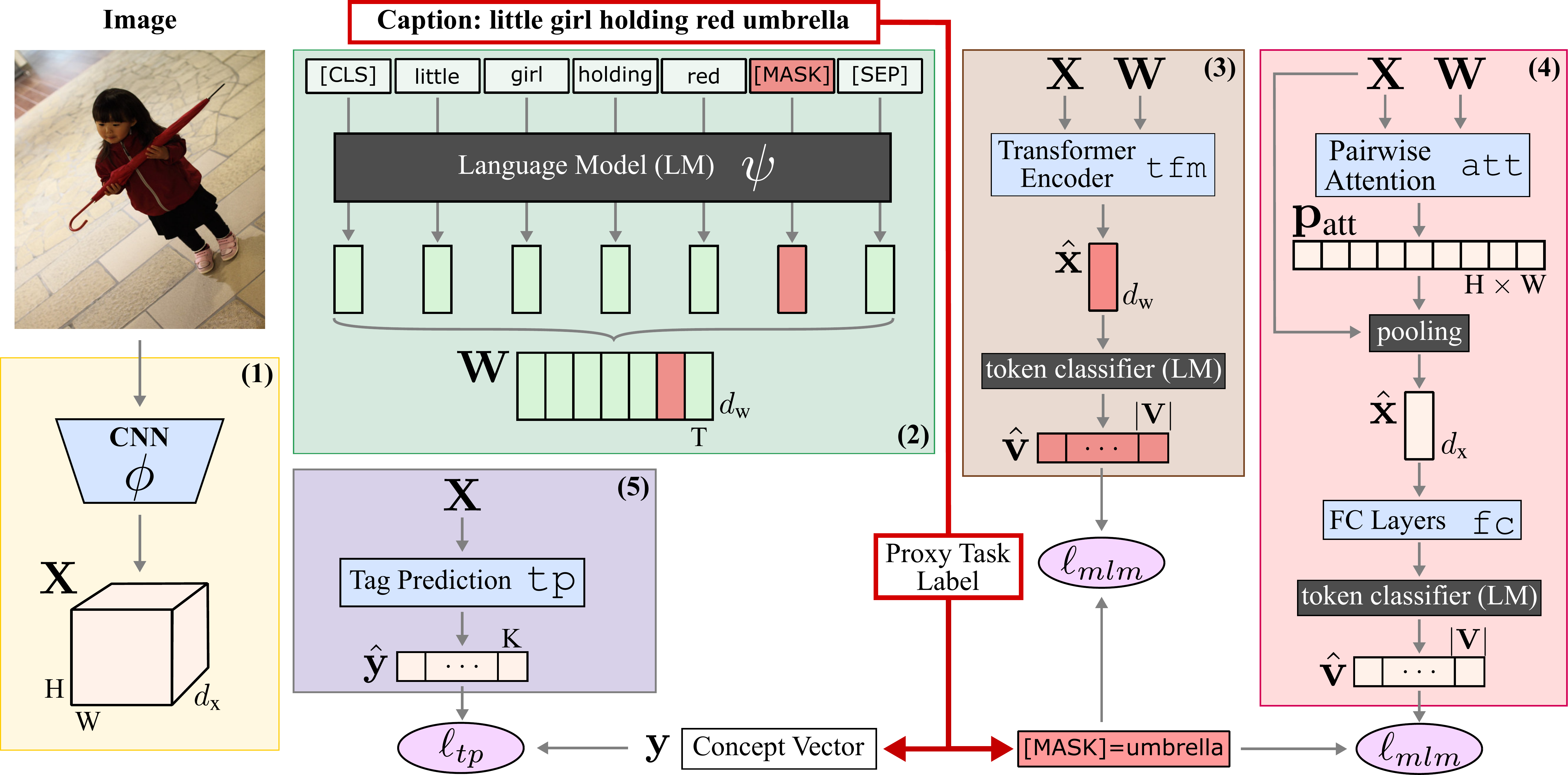}
	\caption{\textbf{Modules used in our models.}
             {\bf (1)} a CNN to extract visual features;
             {\bf (2)} a language model to extract token features;
             {\bf (3)}, {\bf (4)} and {\bf (5)} respectively correspond to our \texttt{tfm}, \texttt{att} + \texttt{fc} and \texttt{tp} modules.
             Our \tpstar{}, \icmlmtfm{} and \icmlmattfc{} models combine these modules: {\bf (1)} + {\bf (5)}, {\bf (1)} + {\bf (2)} + {\bf (3)} and {\bf (1)} + {\bf (2)} + {\bf (4)}, respectively.
             Trainable (and frozen) components are colored in {\color{NavyBlue} blue} (and black).
             Only the CNN is used during target task evaluations.
             }
    \label{fig:icmlm}
\end{figure}

\subsection{Capturing image-level semantics}%
\label{sec:method_tp}%

A straightforward way to build a proxy task given image-caption pairs is to formulate a multi-label image classification problem, where, according to its caption, multiple concepts may appear in an image~\cite{joulin2016learning,quattoni07learning}.
For this setup, we create a label vector $\vec{y} \in \{0, 1 \}^{K}$ for each image $I$ such that $\vec{y}_j = 1$ if concept $o_j$ appears in the image, and $0$ otherwise.
We denote these labels as {\em tags}, and name this task as {\em tag prediction} (TP), illustrated in \fig{fig:icmlm} (modules {\bf (1)} + {\bf (5)}).

One of the contributions of this work is to consider different ways to define concept sets $\C$ from captions.
Ground-truth concept vectors can be easily obtained by considering the most frequent bi-grams~\cite{joulin2016learning} or even n-grams~\cite{li2017ngrams} in captions.
More sophisticated ways to obtain artificial labels include using LDA~\cite{blei2003latent} to discover latent topics in captions~\cite{gomez17self}.
In addition to these existing methods, we look for ways to exploit semantics of tokens in captions.

\mypar{\tppostag{}}
As a first approach, we simply propose to construct label sets by taking into account the {\em part-of-speech} (POS) tags of tokens in captions.
Concretely, we use the off-the-shelf language parser~\cite{spacy} to determine POS tags of tokens in captions and gather three label sets of size $K$, including (i) only nouns, (ii) nouns and adjectives, (iii) nouns, adjectives and verbs.
These three label sets are used to train three separate \tppostag{} models.

\mypar{\tpcluster{}}
As mentioned above, we believe it would be beneficial to use the structure of the full caption and not just treat it as an orderless set of tokens as the previously proposed \tppostag.
To this end, we use the pretrained \bert{}~\cite{devlin2019bert} model to extract sentence-level caption representations.
We do this by feeding the caption into \bert{} and taking the representation for the [CLS] token, which is used as a special token to encode sentence-level text representations in \bert{}.
Then, we cluster the sentence-level representations of all captions using the \kmeans{} algorithm and apply hard cluster assignment.
This way, the labels are the cluster indices and we train $\mphi$ by learning to predict the cluster assignments of captions from their associated image.
$K$-means learns $K$ cluster centroids $\cent^\star \in \Real{\dw \times \nk}$ in the caption representation space by minimizing:
\begin{equation}
\cent^\star, \{ \vec{y}^{i\star} \}_{i=1}^{N} = \argmin_{
\substack{
\cent \in \Real{\dw \times \nk},\\
\{ \vec{y}^i \in \{0,1\}^{\nk},\,\, \mathbf{1}_{\nk} ^\top \vec{y}^i = 1 \}_{i=1}^{N}}}\quad
\sum_{i=1}^N
\| \mpsi(c^i)_\text{[CLS]} -  \cent \vec{y}^i \|_2^2,
\end{equation}
where $\mpsi(c)_\text{[CLS]}$ and $\vec{y}^\star$ denote the [CLS] representation of the caption $c$ and of the one-hot cluster assignment vector obtained for $c$.
Note that $\vec{y}^\star$ is used as the label for image $I$.
In case there are multiple captions for an image, we simply aggregate the cluster labels of all captions associated to that image.

\mypar{Training \tpstar{} models}
Once we have crafted image labels over a chosen set of concepts (either using POS tags or cluster assignments), following~\cite{mahajan2018ECCV}, we normalize the binary label vectors to sum up to one, \ie~$\vec{y}^\top \mathbf{1}_{\nk} = 1, $ for all samples.
Then we train models by minimizing the categorical cross-entropy:
\begin{equation}~\label{eq:tp_loss}
\mlosstp = - \E_{(I, c) \in \D} \Bigg[ \sum_{k=1}^{\nk} \vec{y}_k \log(p(\hat{\vec{y}}_k \, | \, I)) \Bigg],
\end{equation}
where $p(\hat{\vec{y}}_k \, | \, I) = \frac{\exp(\hat{\vec{y}}_k)}{\sum_j \exp(\hat{\vec{y}}_j)}$, $\hat{\vec{y}}_k = \mtp(\mphi(I))_k$, and $\mtp: \Real{\hx \times \wx \times \dx} \rightarrow \Real{\nk}$ is a parametric function performing tag predictions.

\subsection{Capturing localized semantics}%
\label{sec:method_icmlm}%
The previous section presents a cluster prediction task where the structure of the sentence is leveraged through the use of the [CLS] output of the pretrained \bert{}.
Yet, this has a major limitation: token-level details may largely be ignored especially when captions are long~\cite{bahdanau2015neural}.
Our experiments also support this argument, \ie~\tpcluster\ performs on par with or worse than \tppostag.
To address this issue, we propose a second learning protocol that learns to explicitly {\em relate} individual concepts appearing in both an image and its caption.

To this end, we extend the natural language processing task known as Masked Language Model (MLM)~\cite{devlin2019bert} into an {\em image-conditioned} version.
The MLM task trains a language model by masking a subset of the tokens in an input sentence, and then by predicting these masked tokens.
Inspired by this idea, we introduce the Image-Conditioned Masked Language Model (ICMLM) task.
Compared to MLM, we propose to predict masked tokens in a caption by using the visual information computed by $\mphi$.
This way, we learn visual representations that should be informative enough to reconstruct the missing information in captions.

For this task, for each image-caption pair $(I, c)$, we assume that there is at least one concept appearing in the caption $c$.
Since $c$ describes the visual scene in $I$, we assume that concepts appearing in $c$ are observable in $I$ as well.
This allows us to define \icmlm\ as a concept set recognition problem in images.
More precisely, we use the pretrained \bert{} model~\cite{devlin2019bert} as the textual embedding function $\mpsi$ and define the learning protocol as follows.
First, we segment the caption $c$ into a sequence of tokens $(t_{1}, \ldots, t_{T})$, and mask one of the tokens $t_{m}$, which belongs to the concept set.
Masking is simply done by replacing the token $t_{m}$ with a special token reserved for this operation, for instance \bert{}~\cite{devlin2019bert} uses ``[MASK]''.
Then, {\em contextualized} representations of the tokens are computed as $\W = \mpsi((t_{1},\ldots,t_{T}))$.
Meanwhile, the visual representation of the image $I$ is computed by $\mphi(I) = \X$.
Since our goal is to predict the masked token by using both visual and textual representations, we need to merge them.
A naive way to accomplish that is to (i) pool the representations of each modality into a global vector, (ii) aggregate (\ie~concatenate) these vectors, (iii) use the resulting vector to predict the label of the masked token.
However, the representations obtained in this way could only focus on the global semantics, and the local information for both modalities might be lost during the pooling stage.
To address this concern, we describe two possible designs for ICMLM relying on individual visual (in the spatial grid) and textual (in the sequence) features.

\mypar{\icmlmtfm{}}
Here, we contextualize token representations among visual ones by fusing them in a data-driven manner (similar to~\cite{lu2019vilbert}).
Concretely, we spatially flatten and project $\X$ to the token embedding space,
concatenate it with $\W$ and apply a transformer encoder module~\cite{Vaswani2017AttentionIA}, \tfm{}, on top of the stacked representations.
Finally, as done in \bert{}~\cite{devlin2019bert}, the label of the masked token $t_{m}$ can be predicted by feeding the representation of the {\em transformed} masked token into the pretrained token classification layer of \bert{}.
We call this ICMLM flavor \icmlmtfm (modules {\bf (1)} + {\bf (2)} + {\bf (3)} in \fig{fig:icmlm}).

\mypar{\icmlmattfc{}}
Transformer networks employ a self-attention mechanism with respect to their inputs.
Therefore they can learn the pairwise relationships of both the visual and the textual representations.
This allows them, for instance, to fuse different domains quite effectively~\cite{lu2019vilbert,sun2019vidbert}.
We also verify this powerful aspect of the transformers in our experiments, \eg~even a single-layered transformer network is enough to perform significantly well at predicting masked tokens on the MS-COCO dataset~\cite{lin14coco}. 
However, the rest of the caption is already a powerful cue to predict the masked token and this transformer-based architecture might rely too much on the text, potentially leading to weaker visual representations.
As an alternative, we propose to predict the label of the masked token by using the visual features alone.
Since the masked token is a concept that we want to recognize in the image, we divide the prediction problem into two sub-problems: localizing the concept in the image and predicting its label.
To do that we define two additional trainable modules: \att\ and \fc\ modules that we describe in detail below.
This ICMLM flavor is referred to as \icmlmattfc{} (modules {\bf (1)} + {\bf (2)} + {\bf (4)} in Fig.~\ref{fig:icmlm}).

The goal of the \att\ module is to create a 2D attention map on the spatial grid of the visual feature tensor $\X$ such that high energy values correspond to the location of the concept masked in the caption $c$.
It takes as input the spatially-flattened visual features $\X \in \Real{\hx \cdot \wx \times \dx}$ and the textual features $\W$.
First, $\X$ and $\W$ are mapped to a common $\dz$-dimensional space and then pairwise attention scores between visual and textual vectors are computed:
\begin{equation}~\label{eq:att_norm}
\tilde{\X} = \lfloor \mnorm(\X \Sigma_x) \rfloor_+ ,
\hspace{1cm}
\tilde{\W} = \lfloor \mnorm(\W \Sigma_w) \rfloor_+ ,
\hspace{1cm}
\vec{S} =  \frac{\tilde{\X} {\tilde{\W}}^\top}{\sqrt{\dz}},
\end{equation}
where $\Sigma_x \in \Real{\dx \times \dz}$ and $\Sigma_w \in \Real{\dw \times \dz}$ are parameters to learn, $\mnorm$ is LayerNorm~\cite{ba16layer} and $\lfloor . \rfloor_+$ is ReLU operator.
Note that $\vec{S}_{i,j}$ denotes the attention of visual vector $i$ (a particular location in the flattened spatial-grid of the image) to textual vector $j$ (a particular token in the caption).
To be able to suppress attention scores of vague tokens such as ``about'' or ``through'', we compute {\em soft maximum} of the textual attentions for each visual feature:
\begin{equation}~\label{eq:logsumexp}
\vec{s}_i = \log \sum_{j=1}^{T} \exp\left(\vec{S}_{i, j}\right).
\end{equation}
We note that operations in Eqs.~(\ref{eq:att_norm}) and~(\ref{eq:logsumexp}) are performed for a single attention head and the multi-headed attention mechanism~\cite{Vaswani2017AttentionIA} can easily be adopted by learning a weighted averaging layer:
$ \vec{s}_i = \left[ \vec{s}_i^1 | \cdots | \vec{s}_i^H \right] \Sigma_h + b_h $,
where $\Sigma_h \in \Real{H}$ and $b_h \in \Real{}$ are the parameters of the averaging layer, $\vec{s}_i^h$ is the aggregated textual attention score for the $i$\textsuperscript{th} visual feature coming from the $h$\textsuperscript{th} attention head, and $[.|.]$ denotes concatenation.
Finally, attention probabilities are obtained by applying softmax, and used to pool $\X$ into a single visual feature $\hat{\vec{x}}$:
\begin{equation}
{\vec{\matt}}_i = \frac{\exp(\vec{s}_i)}{\sum_{j=1}^{\hx \cdot \wx} \exp(\vec{s}_j)},
\hspace{1cm}
\hat{\vec{x}} = \X^\top \vec{\matt},
\end{equation}
where $\vec{\matt} \in [0, 1]^{\hx \cdot \wx}$ such that $\vec{\matt}^\top \mathbf{1}_{\hx \cdot \wx} = 1$.

After localizing the concept of interest in image $I$ by means of pooling $\X$ into $\hat{\vec{x}}$, we feed $\hat{\vec{x}}$ into the \fc{} module, which consists in a sequence of fully-connected layers, each composed of linear transformation, LayerNorm and ReLU operator.
Finally, we map the output of the \fc{} module to the \bert{}'s token vocabulary $\V$  and compute prediction probabilities as follows:
\begin{equation} \label{eq:probbert}
p_\V\left(k | I, c, t_{m}\right) = \frac{ \exp(\hat{\vec{v}}_{k}) }{ \sum_j \exp(\hat{\vec{v}}_{j})},
\end{equation}
where $\hat{\vec{v}}_{k} = \text{\texttt{fc}}(\hat{\vec{x}})^\top \V_k$ and $\V_k \in \dw$ are the prediction score and the pretrained distributed representation of the $k$\textsuperscript{th} token in the pretrained candidate lexicon of \bert{}.
As we compute dot-products between post-processed $\hat{\vec{x}}$ and the pretrained representations of the tokens in \bert{}'s vocabulary, it is possible to leverage the structure in \bert{}'s hidden representation space.
Indeed, we observe that such probability estimation of a candidate token is more effective than learning a fully connected layer which projects $\text{\texttt{fc}}(\hat{\vec{x}})$ onto the vocabulary.

\mypar{Training \icmlmstar{} models}
To train both model flavors, for each masked token $t_m$ in all ($I$, $c$) pairs in $\D$, we minimize the cross-entropy loss between the probability distribution over the \bert{}'s vocabulary as computed in Eq.~(\ref{eq:probbert}) and the label of the masked token $t_m$ (index of $t_m$ in $\V$):
\begin{equation}~\label{eq:mlm_loss}
\mlossmlm = - \E_{(I, c) \in \D} \left[ \E_{t_m \in c} \Big[ \log ( p_\V (k | I, c, t_{m} ) ) \Big] \right],
\end{equation}
where $k$ is the index of $t_m$ in \bert{}'s vocabulary.
The expectation over captions implies that there can be multiple concepts in a caption and we can mask and predict each of them separately.
For \icmlmtfm, $\hat{\vec{x}}$ is computed by the \tfm\ module, and it corresponds to the representation of the masked token.
For \icmlmattfc, $\hat{\vec{x}}$ corresponds to the output from the \fc\ module.

We also note that $\mlosstp$ and $\mlossmlm$ are complementary, enforcing $\mphi$ to focus on global and local semantics in images, respectively.
Therefore, in both \icmlmattfc\ and \icmlmtfm\ we minimize the weighted combination of $\mlosstp$ and $\mlossmlm$:
\begin{equation}~\label{eq:icmlm_loss}
\mlossicmlm =  \mlossmlm + \lambda \mlosstp .
\end{equation}

{
\newcommand{\mcol}[3]{\multicolumn{#1}{#2}{#3}}
\setlength{\tabcolsep}{0.05cm}
\begin{table}[t!]
    \scriptsize
	\centering
	\caption{{\bf \proxy{Proxy} \vs~\target{target} task performances.}
	          We report top-1 and top-5 masked token prediction scores (as proxy, on VG and COCO) and mAP scores obtained using features from various layers (as target, on VOC-07), on validation sets.\
	          T-1/5: top-1/5 scores,
              C-$\star$: conv. layer from which features are extracted.
	          }%
	\label{tab:icmlm_ablation_vgg16}
    \adjustbox{width=1.0\linewidth}{%
	\begin{tabular}{l | cc cccc | cc cccc}
		\toprule
                          & \mcol{3}{c}{\proxy{Proxy}}                  & \mcol{3}{c|}{\target{Target}}                 & \mcol{3}{c}{\proxy{Proxy}}           & \mcol{3}{c}{\target{Target}}  \\
		Method            & \proxy{Dataset} & \proxy{T-1} & \proxy{T-5} & \target{C-11} & \target{C-12} & \target{C-13}  & \proxy{Dataset} & \proxy{T-1} & \proxy{T-5} & \target{C-11} & \target{C-12} & \target{C-13} \\
		\midrule
		\bert\            & VG              & 17.4        & 36.9        & --            & --            & --             & COCO            & 25.7        & 40.3        & --            & --            & --         \\
		\icmlmtfm\        & VG              & {\bf 49.7}  & {\bf 79.2}  & 71.3          & 75.8          & 80.5           & COCO            & {\bf 70.3}  & {\bf 91.5}  & 70.2          & 74.2          & 77.5       \\
		\icmlmattfc\      & VG              & 41.1        & 71.3        & {\bf 73.7}    & {\bf 78.7}    & {\bf 83.1}     & COCO            & 59.4        & 83.4        & {\bf 72.3}    & {\bf 77.5}    & {\bf 82.2} \\
		\bottomrule
	\end{tabular}
	}
\end{table}
}

\section{Experiments}%
\label{sec:exps}%

This section evaluates
\begin{enumerate*}[(i),font=\bfseries]
    \item how the performance on the masked language modeling (MLM) proxy task translates to target tasks (\sect{sec:exp_mlm_ablation}),
    \item how several types of supervision associated to a set of images (\ie full, weak and self-supervision) compare to each other (\sect{sec:exp_fws_comp}),
    \item if the gains of \icmlmstar{} models are consistent across backbone architectures (\sect{sec:exp_resnet50}),
    \item if \icmlmstar{} models attend to relevant regions in images (Figs \ref{fig:front} and \ref{fig:r50_vis_coco}).
\end{enumerate*}
First, we introduce our experimental setup (remaining details are in the
supplementary material).

\mypar{Datasets}
We train our models on the image-caption pairs of either the 2017 split of MS-COCO~\cite{lin14coco} (COCO) or the Visual Genome~\cite{krishna17vg} (VG) datasets.
COCO has 123K images (118K and 5K for train and val) and 5 captions for each image while VG has 108K images (we randomly split 103K and 5K for train and val) and 5.4M captions.
We remove duplicate captions and those with more than 25 or less than 3 tokens.
We construct several concept sets using the captions of COCO or VG, to be used as tags for \tppostag{} and as maskable tokens for \icmlmstar{} models (an ablative study is provided in the supplementary material).
Note that depending on the concept set, the number of tags and the (image, caption, maskable token) triplets vary, therefore, we specify which concept set is used in all \tppostag{} and \icmlmstar{} experiments.

\mypar{Networks}
To be comparable with the state-of-the-art self-supervised learning method DeeperCluster~\cite{caron19unsupervised}, we mainly use VGG16~\cite{simonyan15VGG} backbones.
We also evaluate \icmlmstar{} models using ResNet50~\cite{he2016resnet} in \sect{sec:exp_resnet50}.
Note that \icmlmstar{} models operate on a set of visual tensors, therefore, for \tpstar{} and \icmlmstar{} models we remove the FC layers from VGG16.
To compensate, we use 4-layered CNNs combined with global average pooling and linear layer for tag predictions as $\tp$ modules.
For $\tfm$, $\att$ and $\fc$ modules, we cross-validated the number of hidden layers and attention heads on the validation set of Pascal VOC-07 dataset, and found that 1 hidden layer (in \tfm{} and \fc{}) and 12 attention heads (in \tfm{} and \att{}) works well.
While training \icmlmstar{} models we set $\lambda = 1$ in \eq{eq:icmlm_loss}.

\mypar{Target task}
Once a model is trained, we discard its additional modules used during training (\ie all but $\mphi$) and evaluate $\mphi$ on image classification tasks, to test how well pretrained representations generalize to new tasks.
To do that, following~\cite{caron19unsupervised}, we train linear logistic regression classifiers attached to the last three convolutional layers of the frozen backbones $\mphi$ with SGD updates and data augmentation.
We perform these analyses on the Pascal-VOC07 dataset~\cite{pascal-voc-2007} (VOC) for multi-label classification, and ImageNet-1K (IN-1K)~\cite{deng2009imagenet} and Places-205~\cite{zhou2014learning} datasets for large-scale categorization, using the publicly available code of~\cite{caron19unsupervised} with slight modifications: We apply heavier data augmentations~\cite{chen20simclr} and train the classifiers for more iterations, which we found useful in our evaluations.

\mypar{Additional \tpstar{} models}
We note that the TP model defined in \sect{sec:method_tp} can be used for predicting any type of image tags, with slight modifications.
We use it to predict topics as proposed in~\cite{gomez17self} and denote this approach as \tplda.
To do so, we only modify \eq{eq:tp_loss} to minimize binary cross-entropy loss instead, where $K$ denotes the number of hidden topics.
Similarly, we denote \tplabel\ as the supervised approach which uses the annotated image labels as tags.

\subsection{Ablative study on the proxy task}%
\label{sec:exp_mlm_ablation}%

We first study the interplay between ICMLM and target tasks.
To do so, we train several \icmlmstar{} models, and monitor their performance on both the proxy and target tasks, \ie~we report masked token prediction (MTP) scores on VG and COCO, and mAP scores on VOC, respectively.
For reference, we also report MTP scores obtained by a single \bert{} model, where masked tokens are predicted using only the remainder of the captions.
In this study, we used the 1K most frequent nouns and adjectives in the captions as maskable tokens.

\noindent {\bf{Results}} are shown in \tab{tab:icmlm_ablation_vgg16}.
We observe that \icmlmstar{} models significantly improve MTP scores compared to \bert{} model, showing that visual cues are useful for MLM tasks.
Moreover, \icmlmtfm{} is better than \icmlmattfc{} on the proxy task, indicating that blending visual and textual cues, which is effectively done by the \tfm{} module,  is beneficial for MLM.
However, \icmlmattfc{} generalizes better than \icmlmtfm{} to VOC.
We believe that, as \icmlmattfc{} predicts masked tokens using visual cues only, it learns semantic concepts from the given training set better than \icmlmtfm{}.
A similar study which uses ResNet50 backbones~\cite{he2016resnet} leads to similar observations (see the supplementary material).

\subsection{Comparison of fully-, weakly- and self-supervised methods}%
\label{sec:exp_fws_comp}%

Next, we compare the visual representations learned by different state-of-the-art fully-, weakly- and self-supervised learning (SSL) models.
We do this by training the models explained below on COCO or VG, then using their backbones $\mphi$ to perform the target tasks, \ie image classification on VOC, IN-1K and Places-205.

{
\newcommand{\ft}[1]{{\color{red} #1}}
\newcommand{\sd}[1]{{\color{orange} #1}}
\setlength{\tabcolsep}{0.05cm}
\begin{table}[ht!]
\scriptsize
\centering
\caption{
{\bf Fully-, weakly- and self-supervised methods} trained with VGG16 backbones.
We report mAP on VOC and top-1 on IN-1K and Places.
For VOC, we report the mean of 5 runs (std. $\leq 0.2$).
We use pretrained models for ImageNet and DeeperCluster, and train other models from scratch.
\#I: number of images in training sets.
$\text{C-}\star$: Conv. layer from which features are extracted.
\ft{Red} and \sd{orange} numbers denote the first and second best numbers in columns.
\supervised{Blue} numbers are not transfer tasks (\ie they use the same dataset for proxy/target).
}
\adjustbox{width=1.0\linewidth}{%
\begin{tabular}{l | ccll | ccc | ccc | ccc}
\toprule
& \multicolumn{4}{c|}{ \proxy{\em Proxy tasks} } & \multicolumn{9}{c }{ \target{\em Target tasks}} \\
& \multicolumn{4}{c|}{  } & \multicolumn{3}{c |}{\target{VOC}} & \multicolumn{3}{c |}{\target{IN-1K}} & \multicolumn{3}{c}{\target{Places}} \\
{\bf Method}         & {\proxy{Dataset}} & \multicolumn{2}{c}{\proxy{Supervision}}  & {\proxy{\# I}} & C-11 & C-12 & C-13 & C-11 & C-12 & C-13 & C-11 & C-12 & C-13 \\
\midrule
ImageNet               & IN-1K\textsubscript{full} & Labels & 1K classes  & 1.3M & \ft{77.5} & \sd{81.0} & \sd{84.7} & \supervised{59.8} & \supervised{65.7} & \supervised{71.8} & 43.0 & 43.5 & 47.3     \\
$\mathcal{S}$-ImageNet & IN-1K\textsubscript{sub}  & Labels & 1K classes  & 100K & 69.3      & 72.4      & 74.1      & \supervised{50.5} & \supervised{52.5} & \supervised{53.8} & 40.9 & 41.6 & 41.1     \\
$\mathcal{S}$-ImageNet & IN-1K\textsubscript{sub}  & Labels & 100 classes & 100K & 67.4      & 69.6      & 70.5      & 47.4              & 48.4              & 46.3              & 39.3 & 39.3 & 35.8     \\
\tplabel{}             & COCO                      & Labels & 80 classes  & 118K & 72.4      & 76.3      & 79.9      & 50.4              & 50.6              & 49.9              & 44.5 & 45.0 & 44.5     \\
\midrule
DeeperCluster~\cite{caron19unsupervised} & YFCC & Self & - & 96M  & 71.4 & 73.3 & 73.1 & 48.0 & 48.8 & 45.1 & 43.1 & 44.1 & 41.0 \\
RotNet~\cite{gidaris2018ICLR}            & COCO & Self & - & 118K & 60.3 & 61.1 & 58.6 & 41.8 & 40.1 & 33.3 & 39.5 & 38.4 & 34.7 \\
RotNet~\cite{gidaris2018ICLR}            & VG   & Self & - & 103K & 59.9 & 60.9 & 59.2 & 39.5 & 38.4 & 34.7 & 39.7 & 38.9 & 34.9 \\
\midrule
\tplda~\cite{gomez17self}   & COCO & Text & 40 topics     & 118K & 70.6 & 73.9 & 76.3 & 48.7      & 48.4      & 46.7      & 43.7 & 44.1      & 43.0 \\
\tpcluster~(\em Ours)       & COCO & Text & 1K clusters   & 118K & 71.5 & 74.5 & 77.0 & 49.5      & 49.8      & 48.1      & 44.1 & 44.6      & 43.7 \\
\tpcluster~(\em Ours)       & COCO & Text & 10K clusters  & 118K & 72.1 & 75.0 & 77.2 & 50.2      & 50.3      & 48.7      & 45.1 & 45.3      & 44.2 \\
\tppostag~(\em Ours)        & COCO & Text & 1K tokens     & 118K & 73.3 & 76.4 & 79.3 & 50.6      & 51.1      & 50.0      & 45.9 & 46.5      & 45.8 \\
\tppostag~(\em Ours)        & COCO & Text & 10K tokens    & 118K & 73.6 & 77.0 & 79.4 & 51.2      & 51.7      & 50.5      & 46.1 & 47.0      & 46.1 \\
\icmlmtfm~(\em Ours)        & COCO & Text & sentences     & 118K & 74.8 & 77.8 & 80.5 & 52.0      & \sd{52.0} & \sd{50.8} & 46.8 & 47.3      & 46.2 \\
\icmlmattfc~(\em Ours)      & COCO & Text & sentences     & 118K & 75.4 & 79.1 & 82.5 & \sd{52.2} & \ft{52.2} & 49.4      & 46.4 & 47.0      & 44.6 \\
\midrule
\tplda~\cite{gomez17self}   & VG   & Text & 40 topics     & 103K & 71.5      & 74.6      & 77.7      & 49.3      & 49.2      & 47.8      & 44.4      & 44.9      & 44.0     \\
\tpcluster~(\em Ours)       & VG   & Text & 1K clusters   & 103K & 73.0      & 76.2      & 79.4      & 50.0      & 49.8      & 47.3      & 45.4      & 45.8      & 44.5     \\
\tpcluster~(\em Ours)       & VG   & Text & 10K clusters  & 103K & 73.9      & 77.8      & 81.3      & 50.8      & 50.7      & 48.5      & 46.2      & 46.9      & 45.6     \\
\tppostag~(\em Ours)        & VG   & Text & 1K tokens     & 103K & 72.9      & 76.4      & 79.6      & 49.9      & 49.8      & 49.1      & 46.0      & 46.5      & 46.4     \\
\tppostag~(\em Ours)        & VG   & Text & 10K tokens    & 103K & 73.5      & 76.9      & 80.1      & 50.9      & 51.3      & 50.0      & 46.1      & 46.7      & 46.7     \\
\icmlmtfm~(\em Ours)        & VG   & Text & sentences     & 103K & 75.5      & 79.3      & 82.6      & \ft{52.4} & \ft{52.2} & \ft{51.1} & \sd{47.3} & \sd{47.8} & \sd{47.5}     \\
\icmlmattfc~(\em Ours)      & VG   & Text & sentences     & 103K & \sd{76.9} & \ft{81.2} & \ft{85.0} & \sd{52.2} & \ft{52.2} & 47.8      & \ft{47.4} & \ft{47.9} & \ft{47.7}\\
\bottomrule
\end{tabular}
}
\label{tab:vgg16_lineval}
\end{table}
}

\mypar{Supervised}
For reference, we report the results obtained by three supervised classifiers trained on different subsets of IN-1K:
\begin{enumerate*}[(i),font=\bfseries]
    \item ``ImageNet'' on the full IN-1K,
    \item ``$\mathcal{S}$-ImageNet with 1K classes'' on randomly-sampled 100 images per class,
    \item ``$\mathcal{S}$-ImageNet with 100 classes'' on 1K images for each of 100 randomly sampled classes.
\end{enumerate*}
The latter two contain 100K images each \ie~the same order of magnitude as COCO or VG.
For the models trained on these three subsets, we repeat the sampling 4 times and report their mean target task results.
We also report \tplabel{} which is trained to predict ground-truth labels.

\mypar{Weakly-supervised}
We compare \tplda{}, \tpcluster{}, \tppostag{} and \icmlmstar{} methods, for which image-level tags are extracted from the captions of COCO or VG.
For \tplda{} we use the publicly-available code of~\cite{gomez17self} to find 40 latent topics among all captions (the number of topics was validated on the validation set of VOC).
Then, probabilities over caption topics define the tag labels for each image.
For \tpcluster{}, we cluster the captions (finding 1K or 10K clusters) and assign the cluster IDs of the captions associated to images as their tag labels.
For \tppostag{}, the tag labels are the most frequent 1K or 10K nouns, adjectives and verbs in the captions.
For \icmlmstar{} models the maskable tokens are the most frequent 1K nouns, adjectives and verbs in the captions.

\mypar{Self-supervised}
For reference, we also provide results for two self-supervised approaches: RotNet~\cite{gidaris2018ICLR} and DeeperCluster~\cite{caron19unsupervised}.
We train RotNet models from scratch on COCO or VG. For DeeperCluster, we use a model pretrained on the large-scale YFCC-100M dataset~\cite{thomee15yfcc100m} (96M images).

\myparnodot{Results} are reported in \tab{tab:vgg16_lineval}.
We observe the following.
\begin{enumerate*}[(i),font=\bfseries]
    \item We see that the good results of ``ImageNet'' are mostly due to its scale.
    Reducing it to 100K images, either by reducing the number of classes or the number of images per class significantly hurt the performance.
    Similarly, the supervised \tplabel{}, which uses an order of magnitude fewer categories and images performs far worse than ImageNet.
    \item The proposed \tpcluster{} outperforms the current state of the art for training with captions, \tplda~\cite{gomez17self}, for all three datasets.
    Exploiting both the structure and the semantics of captions with the \bert{}
    language model, it improves over a topic model.
    However, we see that \tpcluster{} performs on par with or worse than \tppostag{}, suggesting that the importance of individual tokens might be suppressed in global caption representations.
    This validates our motivation for proposing ICMLM in
    \sect{sec:method_icmlm}: models should leverage both global and local semantics in captions.
    \item We see that both \icmlmtfm{} and \icmlmattfc{} improve over all \tpstar{} baselines by significant margins.
    Moreover, on VOC evaluations \icmlmattfc{} outperforms \icmlmtfm{} while on IN-1K and Places it performs on par with or worse than \icmlmtfm{}.
    Note that we observe a similar outcome with ResNet50 backbones (\sect{sec:exp_resnet50}).
    \item Surprisingly, for VOC and Places-205, at least one ICMLM flavor outperforms the full ImageNet pretrained model which we believe is a significant achievement.
    For IN-1K, such comparison does not make sense as, in this setting, the proxy and the target datasets are the same.
    Training on the target set clearly confers an unfair advantage w.r.t. other approaches.
\end{enumerate*}

{
\setlength{\tabcolsep}{0.07cm}
\begin{table}[t]
    \centering
    \begin{minipage}[t]{0.46\linewidth}
        \centering
        \caption{{\bf Fully- and weakly-supervised methods} trained with ResNet50 backbones.
                 We use the pretrained ImageNet model and train other models from scratch.
                 We report mAP and top-1 obtained by linear SVMs (on VOC) and logistic regression classifiers (on IN-1K) using pre-extracted features (avg. of 5 runs, std. $\leq 0.2$).
                 {\supervised{Blue} numbers are not transfer tasks.}}%
        \label{tab:r50_linclf_coco}
        \adjustbox{width=1.0\linewidth}{%
        \begin{tabular}{l | cc | c | c}
        \toprule
        Model         & \proxy{Dataset} & \proxy{Sup.} & \target{VOC} & \target{IN-1K} \\ 
        \midrule
        ImageNet      & IN-1K           & Labels       & {\bf 87.9}   & \supervised{74.7} \\
        \tplabel{}    & COCO            & Labels       & 80.2         & 34.0           \\
        \tppostag{}   & COCO            & Text         & 82.6         & 43.9           \\
        \icmlmtfm{}   & COCO            & Text         & 87.3         & {\bf 51.9}     \\
        \icmlmattfc{} & COCO            & Text         & 87.5         & 47.9           \\
        \bottomrule
        \end{tabular}
        }
    \end{minipage}\hfill
    \begin{minipage}[t]{0.52\linewidth}
        \centering
        \captionof{figure}{{\bf Attention maps} for masked tokens produced by \icmlmtfm{} model with ResNet50 backbone trained on COCO {(darker red means stronger attention)}.}%
        \label{fig:r50_vis_coco}
        \includegraphics[width=\linewidth]{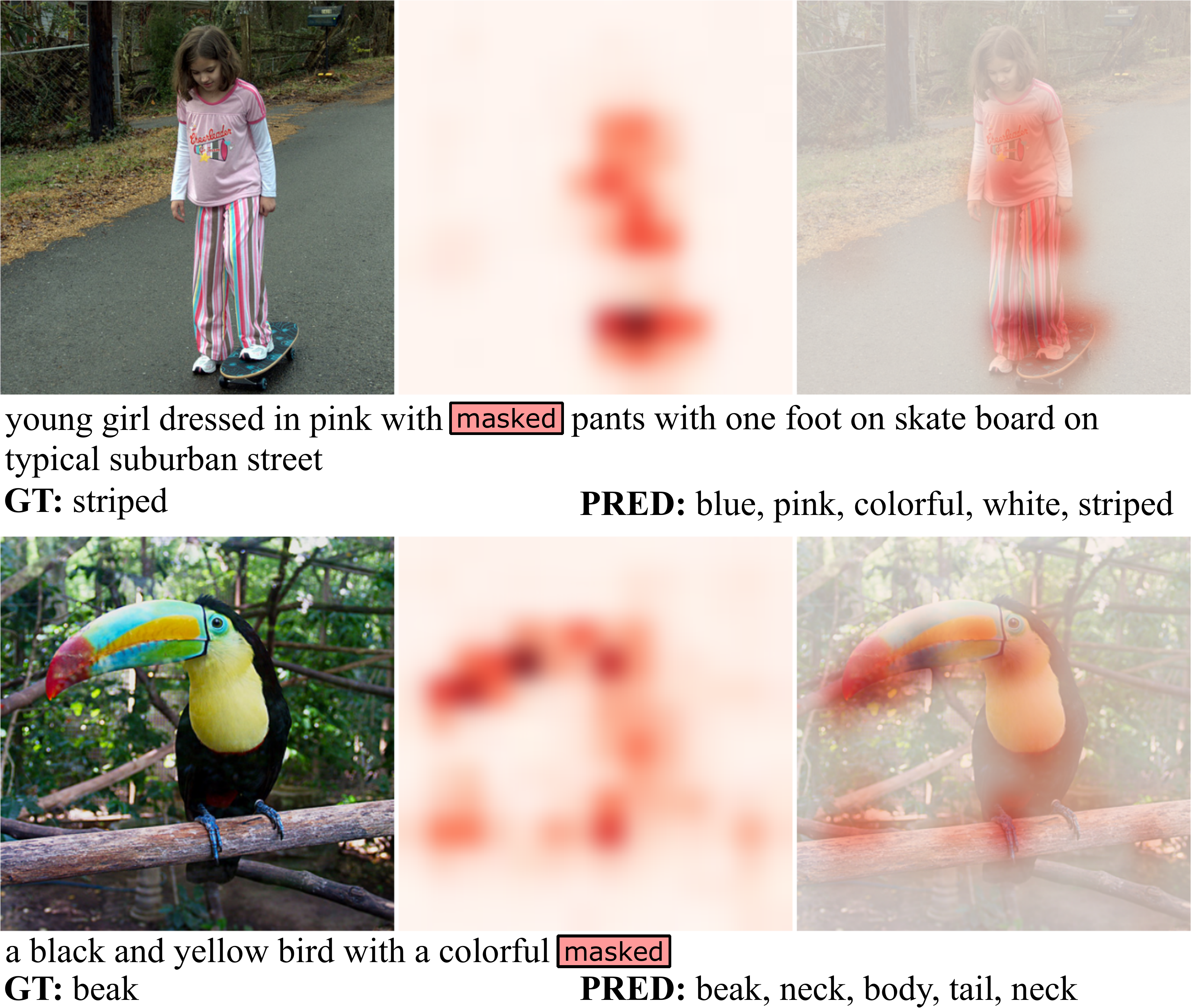}
    \end{minipage}
\end{table}
}

\subsection{Additional results with ResNet50}%
\label{sec:exp_resnet50}%

Some self-supervised proxy tasks might favor certain network architectures (\eg see \cite{kolesnikov2019revisiting}).
This section provides additional results where \icmlmstar{} models use
ResNet50~\cite{he2016resnet}  backbone architectures.
To this end, we train \tplabel{}, \tppostag{} and \icmlmstar{} models on COCO and perform image classification on VOC and IN-1K.
To reduce computational costs, following~\cite{goyal19scaling}, we train linear SVMs (on VOC) and logistic regression classifiers (on IN-1K) using image features pre-extracted from frozen backbones.
Note that ResNet50 is a fully-convolutional network being more expressive compared to VGG16 (thanks to its residual connections and higher number of parameters).
Consequently, in this analysis, we use a 2-layered MLPs as $\tp$ module, a single attention head, and $\lambda = 0.1$ in \eq{eq:icmlm_loss}.
We also move to a bigger concept set for \tppostag{} and \icmlmstar{} models, \ie~the 5K most frequent nouns, adjectives and verbs.

\myparnodot{Results} are shown in \tab{tab:r50_linclf_coco}.
We observe larger improvements of \tppostag{} over \tplabel{} and of \icmlmstar{} over \tppostag{}.
\icmlmstar{} outperforms \tppostag{} by at least $\mathbf{4.7\%}$, $\mathbf{4.0\%}$ and \tplabel{} by at least $\mathbf{7.1\%}$, $\mathbf{13.9\%}$ on VOC and IN-1K.
These results indicate that more complex CNNs are better at suppressing noise in weak labels and at learning cross-modal representations.
Besides, similar to our previous analyses, we see that \icmlmattfc{} learns semantic concepts from the training set slightly better (see the VOC results).
However, \icmlmtfm{} performs better on IN-1K, suggesting that the ResNet50 backbone learns more discriminative features when guided by the same language model.

\mypar{Qualitative results}
Our goal in \icmlm{} is to perform MLM task by {\em looking at} images.
To see if they can attend to relevant parts in images, we visualize {\em attention maps} corresponding to the attention weights of visual features to masked tokens.
Figs \ref{fig:front} and \ref{fig:r50_vis_coco} present such visualizations produced by our \icmlmtfm{} model with ResNet50 backbone trained on COCO.
We see that not only the model is able to detect possible concepts of interest,
it can also understand which concept is asked in the captions (see the supplementary for more visualizations).

\section{Conclusion}%
\label{sec:conc}%
Until recently, carefully collected and manually annotated image sets have provided the most efficient way of learning general purpose visual representations.
To address the annotation cost, weakly-, webly-, and self-supervised learning approaches have traded quality -- a clean supervisory signal -- with quantity, requiring up to hundreds of million images.
Although, in some cases, large quantities of unlabeled data are readily available, processing such large volumes is far from trivial.
In this paper, we seek for a cheaper alternative to ground-truth labels to train visual representations.
First, starting from the observation that captions for images are often easier to collect compared to \eg~fine-grained category annotations, we have defined a new proxy task on image-caption pairs, namely image-conditioned masked language modeling (ICMLM), where image labels are automatically produced thanks to an efficient and effective way of leveraging their captions.
Second, we have proposed a novel approach to tackle this proxy task which produces general purpose visual representations that perform on par with state-of-the-art self-supervised learning approaches on a variety of tasks, using a fraction of the data.
This approach even rivals, on some settings, with a fully supervised pretraining on ImageNet. Such results are particularly relevant for domains where images are scarce but companion text is abundant.

{\small
\bibliographystyle{splncs04}
\bibliography{paper}
}

\ifarxiv
    \clearpage
    \appendix
    \section{Label sets \vs~target task performances}%
\label{sec:exp_ablation_label_sets}%

As we mention in the main paper, for \tppostag{} and \icmlmstar{} models, we can construct multiple concept (or label) sets from captions, \eg~the most frequent $K$ nouns, adjectives or verbs in captions can be used as tags for \tppostag{} and as maskable tokens for \icmlmstar{} models.
In this section, we investigate the impact of learning from such label sets on target task performances.
To do so, we compare learning visual representations using annotated labels of images \vs~tags derived from captions, \ie~\tplabel{} \vs~\tppostag{} with various label sets.

For this analysis, we first train ResNet50 backbones, and then, once a model is trained, we extract image representations from the frozen backbones.
To test generalization capabilities of the representations, we train linear SVMs on VOC~\cite{pascal-voc-2007} and linear logistic regression classifiers on IN-1K~\cite{ilsvrc15}.
Additionally, to understand how effectively models can learn from the training set, we also train linear SVMs on COCO~\cite{lin14coco}.

\myparnodot{Results} are presented in \tab{tab:tp_ablations}.
All \tppostag{} models trained for the ablation improve over \tplabel{}, suggesting that a caption describing an image can provide more comprehensive supervision compared to labeling it with a small set of classes.
It is surprising that gaps are more significant on IN-1K, indicating that a large vocabulary of tags allows backbones to encode more discriminative patterns.
\tppostag{} obtained by using the most frequent 5K nouns, adjectives and verbs in captions improves \tplabel{} by $\mathbf{2.4\%}$, $\mathbf{9.9\%}$ and $\mathbf{2.0\%}$ on VOC, IN-1K and COCO.
In Sec.~4.3 of the main paper, we report results of \tppostag{} and \icmlmstar{} models trained with this label set.

\begin{table}[t]
    \centering
    \scriptsize
    \caption{{\bf \proxy{Label sets} \vs~\target{target} task performances} of \tpstar{} models trained on COCO using ResNet-50 backbones.
             We report mAP (and top-1) scores obtained with linear SVMs on VOC and COCO (and logistic regression classifiers on IN-1K).
             NN, ADJ, VB denote that nouns, adjectives and verbs are present in a label set.
             In parantheses are the number of concepts (\eg classes) in the
             label sets.
             \supervised{Blue} numbers are not transfer tasks.
             }%
    \label{tab:tp_ablations}
    \begin{tabular}{l ccc || l c}
    \toprule
    \proxy{Label Set}          & \target{VOC} & \target{IN-1K} & \target{COCO}            & \proxy{Label Set}    & \target{VOC} \\
    \midrule                                                                             
    GT Labels (\tplabel{}, 80) & 80.2         & 34.0           & \supervised{73.5}        & NN + ADJ + VB (1K)   & 81.4 \\
    NN            (5K)         & 81.8         & 43.9           & \supervised{75.3}        & NN + ADJ + VB (2.5K) & 82.1 \\ 
    NN + ADJ      (5K)         & 82.3         & {\bf 44.5}     & \supervised{{\bf 75.5}}  & NN + ADJ + VB (5K)   & {\bf 82.6} \\ 
    NN + ADJ + VB (5K)         & {\bf 82.6}   & 43.9           & \supervised{{\bf 75.5}}  & NN + ADJ + VB (10K)  & 81.9 \\ 
    \bottomrule
    \end{tabular}
\end{table}

{
\begin{table}[t]
    \centering
    \scriptsize
    \caption{{\bf \proxy{\icmlm{}} \vs~\target{target} task performances.}
             We train \icmlmstar{} models with different numbers of hidden layers (\#L) and attention heads (\#H) on COCO using ResNet-50 backbones and compare them on \proxy{proxy} and \target{target} tasks.
             While training \icmlmstar{} models we set $\lambda = 0$ in Eq.~8 of the main paper.
             For the \proxy{proxy} task, we report top-1 MTP scores on COCO; for the \target{target} tasks see the caption of \tab{tab:tp_ablations}.
             \bert{} alone achieves $25.7\%$ on the \proxy{proxy} task.
             \supervised{Blue} numbers are not transfer tasks.
             }%
    \label{tab:icmlm_ablations}
    \begin{tabular}{cc | cccc | cccc}
    \toprule
              &          & \multicolumn{4}{c|}{\icmlmtfm{}} & \multicolumn{4}{c}{\icmlmattfc{}} \\
    \#L       & \#H      & \proxy{Proxy} & \target{VOC}  & \target{IN-1K} & \target{COCO}           & \proxy{Proxy} & \target{VOC} & \target{IN-1K} & \target{COCO}           \\
    \midrule
         1    &    1     &  65.2         & 85.7          & 50.6           & \supervised{{\bf 77.6}} & 58.5          & {\bf 86.8}   & 47.2           & \supervised{{\bf 78.9}} \\
         1    &    4     &  66.1         & 85.3          & 50.7           & \supervised{77.5}       & 59.4          & 86.7         & 46.8           & \supervised{{\bf 78.9}} \\
         1    &    12    &  66.5         & 85.5          & 50.4           & \supervised{77.2}       & 59.5          & 86.6         & 47.3           & \supervised{{\bf 78.9}} \\
         2    &    1     &  66.7         & 85.0          & 46.6           & \supervised{76.2}       & 59.5          & 86.4         & 48.1           & \supervised{78.8}       \\
         2    &    4     &  67.1         & 85.0          & 46.7           & \supervised{76.3}       & 60.2          & 86.3         & 48.5           & \supervised{78.6}       \\
         2    &    12    &  {\bf 67.5}   & 84.8          & 46.6           & \supervised{76.1}       & {\bf 60.4}    & 86.3         & {\bf 48.7}     & \supervised{78.5}       \\
    \bottomrule
    \end{tabular}
\end{table}
}

\section{\icmlm{} \vs~target task performances}%
\label{sec:exp_ablation_icmlm}%
This section extends the analysis reported in Sec.~4.1 of the main paper, \ie~we study how the masked language modeling (MLM) performance (the proxy task) translates to target tasks. This time we use ResNet50 backbones instead of VGG16 (as in Sec.~4.1 of the main paper).
To do so, we train \icmlmtfm{} (and \icmlmattfc{}) models with different numbers of hidden layers and attention heads in \tfm{} (and, in \fc{} and \att{}, respectively) modules, and monitor both proxy and target task results.
While training \icmlmstar{} models, we set $\lambda = 0$ in Eq.~8 of the main paper: for this ablation study the training solely depends on $\mlossmlm$ defined by Eq.~7 in the main paper.
Similar to the previous analysis, we perform target tasks using pre-extracted image features on VOC, IN-1K and COCO.
We also report top-1 masked token prediction (MTP) scores on COCO.

\myparnodot{Results} are reported in \tab{tab:icmlm_ablations}.
We observe that having more hidden layers or attention heads improve the MLM performance at the expense of reduced target task results.
We believe that as the complexity of \tfm{}, \att{} or \fc{} modules increase, they can learn more interconnections between visual and textual cues, and this, in turn, lifts some of the burden of capturing the semantics of the caption off the visual model itself, and leads $\phi$ to learn weaker visual features.
Moreover, similar to the observations we made in Secs.~4.1 and~4.2 of the main paper, \icmlmtfm{} significantly outperforms \icmlmattfc{} on MLM and IN-1K, however, \icmlmattfc{} is slightly better than \icmlmtfm{} on VOC and COCO.
The fact that IN-1K performance of \icmlmattfc{} increases when \fc{} module has two hidden layers also supports the hypothesis that \icmlmattfc{} tends to overfit to the concepts present in the training set (hence it performs better on VOC and COCO).

Comparing \tab{tab:tp_ablations} and \tab{tab:icmlm_ablations}, we see overall that \icmlmstar{} (when \#L and \#H are 1) improves \tppostag{} by at least $\mathbf{3.1\%}$, $\mathbf{3.3\%}$, $\mathbf{2.1\%}$ and \tplabel{} by at least $\mathbf{5.5\%}$, $\mathbf{13.2\%}$, $\mathbf{4.1\%}$ on VOC, IN-1K and COCO.

\mypar{A note for \icmlmstar{} models with a VGG16 backbone}
We tried these settings for VGG16 backbones: one attention head in \icmlmstar{} models and $\lambda = 0.$ (Eq.~8 of the main paper) but this lead to inferior models.
We believe that this is due to the absence of residual connections in the backbone architecture, which leads to overfitting to MLM tasks (a similar behavior is observed in~\cite{kolesnikov2019revisiting} for self-supervised learning methods trained with VGG16 architecture).

\mypar{Importance of $\lambda$ in Eq.~8 of the main paper}
We discuss in Sec.~3 of the main paper that global \vs~localized semantics in images can and should be captured separately.
To this end, in Eq.~8 of the main paper, we propose to optimize a combination of $\mlosstp$ and $\mlossmlm$ losses to effectively train backbones by providing supervision for both global and localized semantics.
In our \icmlmstar{} experiments, we validated the coefficient $\lambda$ combining these loss terms by monitoring the $\mlosstp$ loss on the validation sets of COCO or VG.
We tried three values for $\lambda \in \{0.0, 0.1, 1.0\}$ and found that $\lambda = 0.1$ and $\lambda = 1.0$ minimize $\mlosstp$ loss on the validation sets with ResNet-50 and VGG16 backbones respectively, and moreover improve target task results.
This finding supports our claim that $\mlosstp$ and $\mlossmlm$ loss terms are complementary.

    \section{Zero-shot Object Classification}%
\label{sec:exp_zsl}%

We also extend the analysis in Sec.~4.2 of the main paper on an additional target task, zero-shot image classification, on CUB-200-2011 (CUB)~\cite{wah2011CUB} and Attributes with Animals 2 (AWA2) datasets~\cite{xian2018gbu}.
The CUB dataset contains roughly 12K images for 200 types of {\em fine-grained} bird species defined by 312 different semantic {\em attributes}.
The AWA2 dataset has roughly 38K images for 50 {\em coarse-grained} animals defined by 85 different attributes.
The classes in these datasets are split into two subsets called {\em seen} and {\em unseen} classes.
The goal of these benchmarks is to train a classification model on seen classes in a way that the classification model can effectively be used for both seen and unseen classes.
Using the recently proposed splits~\cite{xian2018gbu}, we have 150 (resp. 40) and 50 (resp. 10) classes in the seen and the unseen splits for CUB (and AWA) datasets.
Image samples from seen classes are divided into training and test sets whereas image samples from unseen classes are solely used for testing purposes.

In this analysis, we take the VGG16 backbones trained by \tpstar{} or \icmlmstar{} models on the MS-COCO~\cite{lin14coco} (COCO) or Visual Genome~\cite{krishna17vg} (VG) datasets.
Similar to what we report in Sec.~4.2 from the paper, using the activations from the last three convolutional layers, we train bilinear score functions~\cite{sariyildiz2019CVPR} that measure the {\em compatibility} between the visual features $\vec{x} \in \Real{m}$ (pooled and flattened to have roughly 9K dimension) and class-level attribute vectors $\vec{a} \in \Real{n}$ ($n$ is 312 for CUB and 85 for AWA).
Concretely, we define the score function as
\begin{equation}
f(\vec{x}, \vec{a}) = \vec{a}^\top (\vec{\Sigma} \vec{x} + \vec{b})
\end{equation}
where $\vec{\Sigma} \in \Real{n \times m}$ and $\vec{b} \in \Real{n}$ are parameters of the score function to be learned.
Using the score function, class predictions are simply made by:
\begin{equation}
\hat{y} = \,\, \argmax_{c \in \mathcal{C}} \,\, f(\vec{x}, \vec{\mathcal{A}}_c),
\end{equation}
where $\vec{\mathcal{A}}_c \in \Real{n}$ denotes the class-level attribute vector for class $c$ and $\mathcal{C}$ is the set of all classes.
We train the score function by minimizing the following:
\begin{equation}
\vec{\Sigma}^\star, \vec{b}^\star = \argmin_{\vec{\Sigma}, \vec{b}} \,\,\, - \E_{(\vec{x}, y) \in \mathcal{D}} \Big[  \log \left( p \left(y | \vec{x} , \vec{\mathcal{A}} \right) \right) \Big],
\end{equation}
where $\mathcal{D}$ is a dataset of feature-label pairs $(\vec{x}, y)$ s.t. $y \in \{ 1, \ldots, \mathcal{C} \}$ and
\begin{equation}
p \left( y = c | \vec{x} , \vec{\mathcal{A}} \right) = \frac{ \exp( f(\vec{x}, \vec{\mathcal{A}}_c) )}{\sum_j \exp(  f(\vec{x}, \vec{\mathcal{A}}_j) )}.
\end{equation}

\newcommand{\mcol}[2]{\multicolumn{#1}{c}{#2}}
\newcommand{\mt}[1]{{T#1}}

\begin{table}[t]
\centering
\scriptsize
\caption{
{\bf Zero-shot object classification} with VGG16 backbones.
We report top-1 accuracies over all classes (seen + unseen) on CUB and AWA2 datasets.
Those are obtained by training a bilinear function between the visual features produced by each of the methods and the class-level attribute vectors.
We report the mean of 5 runs with different seeds (std. $\le 0.3$ for all settings).
{\#I}: The number of images in the training set.
~\cite{kornblith19transfer} shows that transfer learning performance is correlated with the overlap of classes between IN-1K and target task datasets.
The fact that IN-1K contains 59 bird-related classes and the majority of the classes in AWA2 dataset provides ImageNet pretrained models an unfair advantage.
Therefore, we distinguish them with \supervised{blue} numbers.
}
\label{tab:vgg16_zsl}
\begin{tabular}{=l | +c +l +c | +c | +c | +c | +c | +c | +c}
\toprule
                & \multicolumn{3}{c|}{\proxy{Proxy tasks}}             & \mcol{3}{\target{CUB}} & \mcol{3}{\target{AWA2}} \\
\textbf{Method} & \proxy{Dataset} & \proxy{Supervision} & \proxy{\#I}  & C-11 & C-12 & C-13 & C-11 & C-12 & C-13 \\
\midrule
\rowstyle{\color{NavyBlue}}
ImageNet               & IN-1K & 1K classes   & 1.3M & 10.2 & {\bf 19.4} & {\bf 24.4} & 11.4 & {\bf 37.1} & {\bf 38.9} \\
\rowstyle{\color{NavyBlue}}
$\mathcal{S}$-ImageNet & IN-1K & 1K classes   & 100K & 11.6 & 16.1 & 18.3 & 12.7 & 33.2 & 34.9 \\
\rowstyle{\color{NavyBlue}}
$\mathcal{S}$-ImageNet & IN-1K & 100 classes  & 100K & {\bf 12.5} & 14.1 & 15.7 & {\bf 13.1} & 32.0 & 33.3 \\
\midrule
\tplabel{}             & COCO  & 80 classes   & 118K & 11.1 & 11.7 & 11.5 & 31.1 & 32.0 & 32.8 \\
\tpcluster~(\em Ours)  & VG    & 1K clusters  & 103K &  9.8 & 10.3 & 10.3 & 30.3 & 30.8 & 30.6 \\
\tpcluster~(\em Ours)  & VG    & 10K clusters & 103K & 10.3 & 10.7 & 10.4 & 30.9 & 31.6 & 31.9 \\
\tppostag~(\em Ours)   & VG    & 1K tokens    & 103K & 10.6 & 11.1 & 11.5 & 30.8 & 31.7 & 32.3 \\
\tppostag~(\em Ours)   & VG    & 10K tokens   & 103K & 10.4 & 10.9 & 11.3 & 31.0 & 31.9 & 32.4 \\
\icmlmtfm(\em Ours)    & VG    & sentences    & 103K & {\bf 12.5} & {\bf 13.0} & {\bf 13.7} & {\bf 32.2} & 32.8 & 33.1 \\
\icmlmattfc(\em Ours)  & VG    & sentences    & 103K & 12.1 & 12.0 & 10.9 & 31.5 & 32.1 & 31.5 \\
\icmlmtfm(\em Ours)    & COCO  & sentences    & 118K & 12.4 & 12.8 & 13.3 & {\bf 32.2} & {\bf 32.9} & {\bf 33.8} \\
\icmlmattfc(\em Ours)  & COCO  & sentences    & 118K & 11.9 & 12.3 & 12.4 & 31.8 & 32.7 & 33.1 \\
\bottomrule
\end{tabular}
\end{table}

\mypar{Results}
\tab{tab:vgg16_zsl} reports top-1 prediction accuracies among all classes for both datasets.
We make the following observations.
\begin{enumerate}[(i), font=\bfseries]
    \item We see that \icmlmtfm{} model trained on VG significantly improves \tpstar{} models on CUB, \ie~up to $\mathbf{1.4\%}$, $\mathbf{1.3\%}$ and $\mathbf{2.2\%}$ with C-11, C-12 and C-13 features.
          On the other hand, \icmlmtfm{} model trained on COCO improves \tpstar{} models on AWA2 up to $\mathbf{1.1\%}$, $\mathbf{0.9\%}$ and $\mathbf{1.0\%}$ with C-11, C-12 and C-13 features.
          In fact, \icmlmstar{} models tend to perform slightly better on AWA2 (particularly with C-13 evaluations), when they are pretrained on COCO indicating that the concepts in COCO are semantically more similar to the concepts in AWA2.
    \item When trained on VG, the C-13 features learned by \icmlmattfc{} are inferior to \tppostag{}, \ie~the scores drop up to $0.9\%$.
          This implies that the VGG16 backbone trained by \icmlmattfc{} slightly overfits to MLM task.
          However, the opposite is true for the C-11 and C-12 features, suggesting that the network is able to extract richer semantics from the earlier layers.
\end{enumerate}

    \section{Additional qualitative results}%
\label{sec:exp_vis}%

In Figs.~1 and~3 of the main paper, we show attention maps produced by our \icmlmtfm{} model (\tfm{} module contains 1 hidden layer and 1 attention head) with ResNet-50 backbone trained on COCO.
This section provides additional attention maps obtained by the \att{} module in our \icmlmattfc{} model (\fc{} and \att{} modules contain 1 hidden layer and 12 attention heads, respectively) with VGG16 backbone trained on COCO.
These maps are shown in Figs~\ref{fig:attention_1} and~\ref{fig:attention_2}.

\begin{figure*}[t!]
   \includegraphics[width=\linewidth]{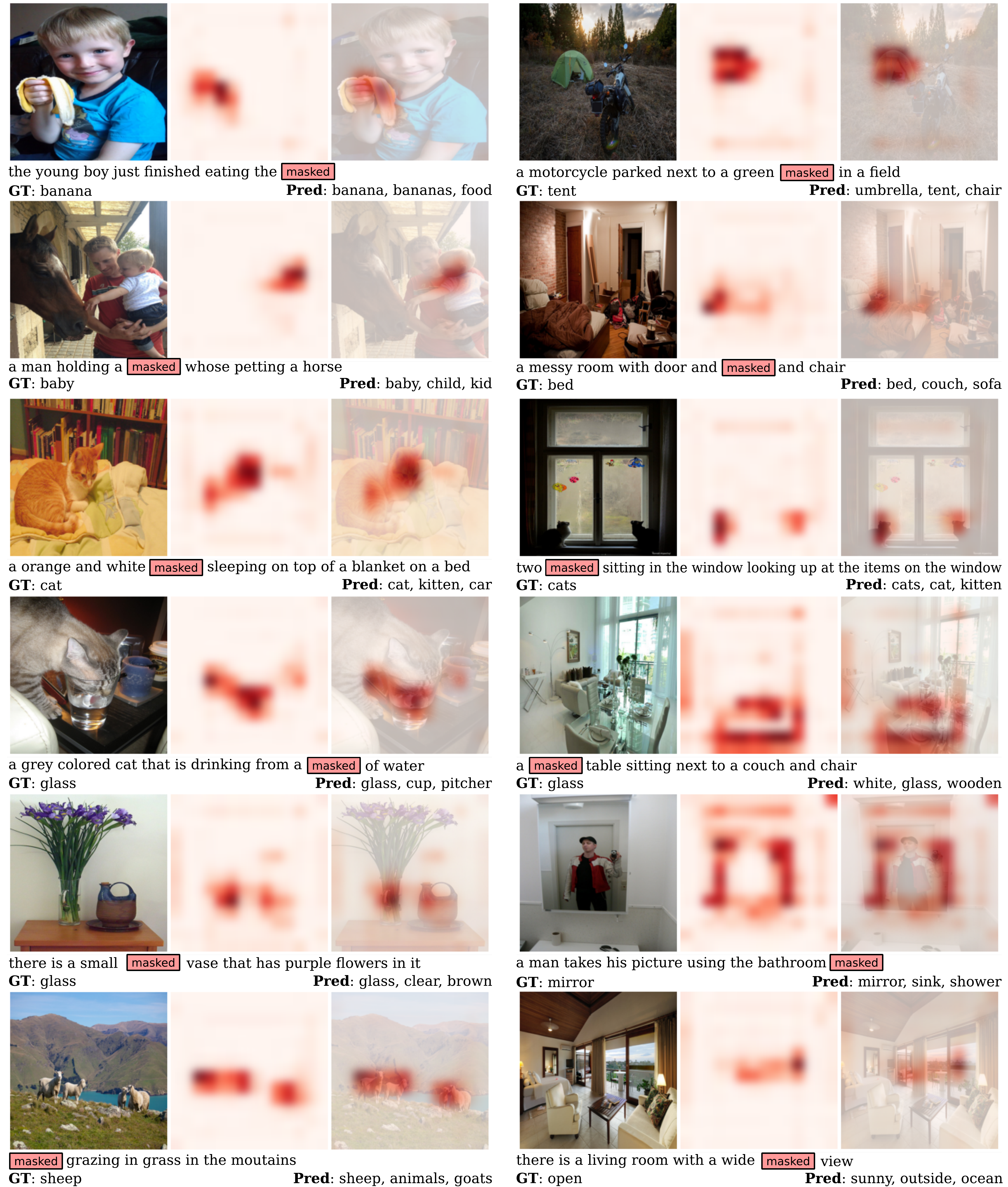}
   \caption{\textbf{Qualitative results.}
            For several image-caption pairs of the validation set of the COCO dataset and for a masked token, we show the ground-truth label (GT) together with the top 3 predictions (Pred) and the attention map generated by our \icmlmattfc{} model with VGG16 backbone.
            The red parts correspond to higher attentions.}
   \label{fig:attention_1}
\end{figure*}

\begin{figure*}[t!]
   \includegraphics[width=\linewidth]{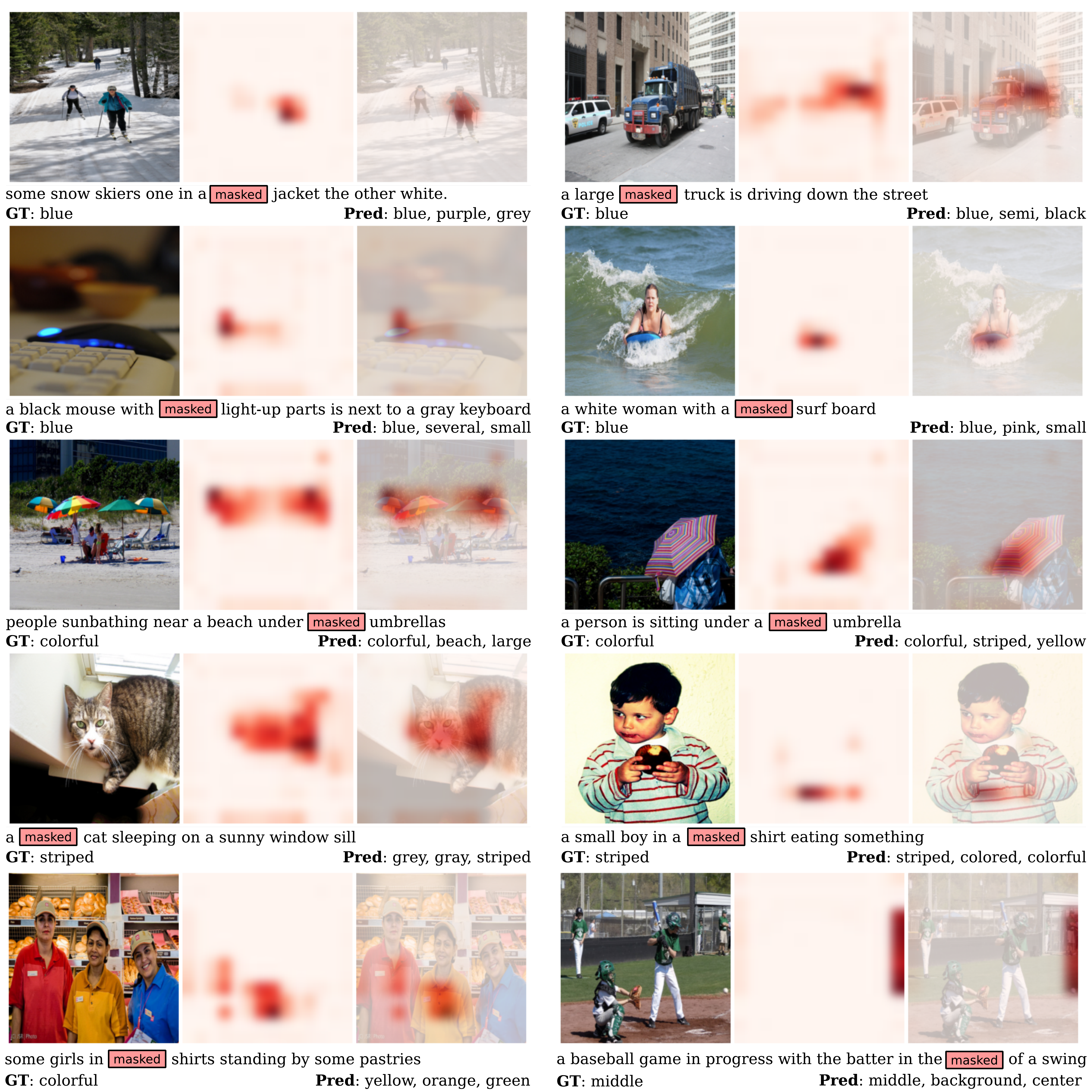}
   \caption{\textbf{Qualitative results.}
            For several image-caption pairs of the validation set of the COCO dataset and for a masked token, we show the ground-truth label (GT) together with the top 3 predictions (Pred) and the attention map generated by our \icmlmattfc{} with VGG16 backbone.
            The red parts correspond to higher attentions.}
   \label{fig:attention_2}
\end{figure*}

First, we see from the figures that the \att{} module can successfully localize object categories that have a clear visual appearance. 
This is the case for instance of the banana, the baby, the cats, or the sheep from Fig.~\ref{fig:attention_1}.  
This is also the case even in cluttered scenes, such as the bed on the second row of Fig.~\ref{fig:attention_1}.

Second, it is interesting to see that even visual concepts that are more abstract than object categories can also be localized, such as the {\em mirror} or {\em glass}. 
In the particular case of the {\em glass} category, the versatility of this concept is successfully captured by our model, covering the drinking glass and the material of the table and of the vase.

Third, the model goes beyond nouns and learns the visual appearance associated to colors or textures.
For instance, the concepts {\em blue}, {\em striped} or {\em colorful} are illustrated in Fig.~\ref{fig:attention_2}. 

Finally, we show some failure cases.
This is often the case for ambiguous concepts whose visual appearance is not properly defined, such as {\em middle} and {\em open} which are respectively illustrated in the bottom right of Fig.~\ref{fig:attention_1} and Fig.~\ref{fig:attention_2}.
In some extreme cases, the attention maps are meaningless, and the masked word prediction relies on the rest of the caption instead.
An other failure case is the bottom left of Fig.~\ref{fig:attention_2} which shows that grouping several concepts (like the different colors of the three shirts) is still way beyond the capacity of the ICMLM model. 

    \section{Transformer network in ICMLM}%
\label{sec:model_tfm}%

This section extends Sec.~3.2 of the main paper and describes in detail the transformer encoder layer~\cite{Vaswani2017AttentionIA} in our \icmlmtfm{} model.

In \icmlmtfm{}, we use the multi-headed attention network proposed in~\cite{Vaswani2017AttentionIA} in order to contextualize the token embeddings computed by \bert{} model, \ie~$\W_i \in \Real{T \times \dw}$, among the visual features mapped to the token embedding space of \bert{}, \ie~$\bar{\X}_i \in \Real{\hx \times \wx \times \dw}$, for the $i$-th data sample.
To do so, in our model, we use 1-layer transformer encoder with 8 attention heads which are computed in parallel.
The transformer encoder takes as input the concatenation of $\bar{\X}_i$ and $\W_i$, \ie~$Z_i = [ \bar{\X}_i; \W_i ] \in \Real{S \times \dw}$, where $S = (\hx \times \wx + T)$ denotes the total number of (visual + textual) tokens.

Each attention head $O_h, h \in {1,\cdots,8}$ in the encoder performs the {\em scaled dot-product attention}~\cite{Vaswani2017AttentionIA} on top of $Z_i$ as follows.
First, 3 linear projections of $Z_i$ are computed:
\begin{equation}
\begin{split}
    K_i^h &= Z_i \Sigma_K^h + b_K^h, \\
    Q_i^h &= Z_i \Sigma_Q^h + b_Q^h, \\
    V_i^h &= Z_i \Sigma_V^h + b_V^h,
\end{split}
\end{equation}
where $K_i^h$, $Q_i^h$ and $V_i^h$ are respectively the keys, queries and values $\in \Real{S \times \dw}$ computed by the attention head $h$.
In this formulation, $\Sigma_K^h$, $\Sigma_Q^h$ and $\Sigma_V^h$ $\in \Real{\dw \times \dw}$ are weight; $b_K^h$, $b_Q^h$ and $b_V^h$ $\in \Real{\dw}$ are bias parameters of the projection layers in $O^h$.
Then the output of each head $O^h (Z_i) \in \Real{S \times \dw}$ is computed using the keys, queries and values defined above:
\begin{equation}
    O^h (Z_i) = \text{softmax}\left( \frac{{K_i^h} {Q_i^h}^\top}{\sqrt{D}} \right) V_i^h.
\end{equation}
 
 Finally all attention heads are merged simply by concatenating the individual head's outputs, and we compute:
 \begin{equation}
     O(Z_i) = \left[ O^1(Z_i) \, | \cdots | \, O^8(Z_i) \right] \Sigma^O + b^O,
 \end{equation}
 where $\Sigma_O \in \Real{8 \times \dw \times \dw}$ and $b^O \in \Real{\dw}$ are learnable parameters, and $[.|.]$ denotes concatenation.
 The output of the multi-headed attention layer is followed by residual connection~\cite{he2016resnet}, dropout~\cite{srivastava14dropout}, LayerNorm~\cite{ba16layer}, ReLU and linear projection layers to obtain the final output of the transformer.

    \section{Implementation details}%
\label{sec:exp_imp}%

This section provides technical details of both training model for proxy tasks and evaluating them on target tasks.

\subsection{Training for proxy tasks}

\mypar{With VGG16 backbones}
We start training VGG16 networks on the Visual Genome (VG) or MS-COCO datasets by solving the rotation prediction task~\cite{gidaris2018ICLR}.
Note that we do not use any of the existing RotNet~\cite{gidaris2018ICLR} pretrained models as they all have processed millions of images.
Contrarily, we want to restrict all the training steps of our pipeline to access only a small dataset of images (103K and 118K training images of VG and COCO respectively).
For that, first, we train separate VGG16 networks on VG or COCO for 120K iterations using RAdam~\cite{liu20radam} with batches of size 128, initial learning rate $1e-3$, weight decay $1e-5$, and the learning rate is decayed by $0.1$ after 100K and 110K iterations.
Once the networks are trained for the rotation prediction task, we remove the fully-connected layers from the networks and fine-tune the CNN backbones by solving the proxy tasks we defined in Secs.~3.1 and~3.2 of the main paper.

We train \tpstar{} models for 100K iterations using RAdam optimizer~\cite{liu20radam} with batches of size 128, initial learning rate 1e-4, weight decay 1e-3, and the learning rate is decayed by 0.1 after 80K and 90K iterations.
For \tpstar{} models, the number of data samples is equal to the number of images in the training sets (103K in VG and 118K in COCO).
The number of unique triplets (image, caption, masked token) that we use during training \icmlm{} models varies from 2.5M to 13M depending on the dataset and the label set used, because we design the triplets in a way that for each (image, caption) pair, there is only one masked token so many triplets are built for a single (image, caption) pair.
To reduce the training time, we train them for 200K iterations using batches of size 896 (distributed over 4 NVIDIA V100 GPUs).
We note that in early \icmlm{} trainings, attention heads (\att{} modules in \icmlmattfc{} and self-attention attention heads in \icmlmtfm{}) produce almost uniform attention distributions over the spatial grid of visual features.
Therefore, in \icmlmattfc{} models, we find that warming up the attention heads for 50K iterations while freezing VGG16 backbones prevents noisy gradients to flow through backbones.

\mypar{With ResNet50 backbones}
We train \tplabel{} and \tppostag{} models from scratch for 100K iterations using SGD with momentum (0.9) optimizer with batches of size 128, initial learning rate 3e-2, weight decay 1e-4, and the learning rate is decayed by a cosine-based schedule.
We initialize ResNet50 backbones in \icmlmstar{} models with pretrained \tppostag{} checkpoints then train \icmlmstar{} models for 500K iterations using the same optimizer configuration except that batch size is 512.

We validate all hyper-parameters and design choices on the validation sets of VG and COCO.
As we note in Sec.~3.2 of the main paper, while training \icmlmstar{} models, we freeze the pretrained \bert{} model available in HuggingFace repository\footnote{\url{https://github.com/huggingface/transformers}}.
We use PyTorch~\cite{paszke19pytorch} and the mixed-precision functionality provided by NVIDIA Apex\footnote{\url{https://github.com/NVIDIA/apex}} to perform all experiments.

\subsection{Evaluation on target tasks}

We follow two different evaluation practices to compare models:
\begin{enumerate}[(i),font=\bfseries]
    \item Probing linear logistic regression classifiers after various layers in VGG16 backbones and training them with SGD updates and data augmentation.
          For this evaluation, we use the publicly-available code of \cite{caron19unsupervised} and slightly modify it such that heavier data augmentation is applied and classifiers are trained for more iterations.
          We will share the training configuration for each setting.
          For the details of the evaluation practice, please refer to the code repository of~\cite{caron19unsupervised}\footnote{\url{https://github.com/facebookresearch/DeeperCluster}}.
    \item Extracting image features from the last convolutional layer of ResNet50 backbones and training linear SVMs and logistic regression classifiers using these pre-extracted features.
\end{enumerate}
Note that in both cases, backbones are frozen.

\mypar{Feature extraction}
To extract image features, we resize images such that their smallest dimension is $224$ pixels, then apply central-crops of size $224 \times 224$.
This gives us $7 \times 7 \times 2048$-dimensional visual tensors output for ResNet-50 backbones.
For training SVMs on VOC and COCO, following~\cite{goyal19scaling}, we apply $2 \times 2$ spatial average pooling and flattening to obtain 8192-dimensional visual features, then $\ell_2$-normalize the features.
However, storing and training classifiers on 8192-dimensional features for the 1.28M images of the IN-1K dataset was computationally challenging.
Therefore, for training logistic regression classifiers on IN-1K, we apply global average pooling and obtain 2048-dimensional visual features.

\mypar{SVM classifiers}
Following the convention of~\cite{goyal19scaling}, we train linear SVMs to evaluate visual representations on the 2007 split of Pascal-VOC and the 2017 split of MS-COCO datasets, in a one-\vs-all manner.
Please refer to~\cite{goyal19scaling} for details in training binary SVMs.
Different from~\cite{goyal19scaling}, we tune the cost parameter of SVMs by sampling 40 cost values log-uniformly between $10^{-5}$ and $10^5$ and find the optimal value by Optuna~\cite{optuna2019}.

\mypar{Logistic regression classifiers}
We train linear logistic regression classifiers by performing SGD updates with momentum $0.9$ and batch size $1024$.
We validate the learning rate and weight decay hyper-parameters using Optuna~\cite{optuna2019} over 25 trials.
We log-uniformly sample learning rates between $10^{-1}$ and $10^{2}$, and apply cosine-based learning rate annealing, whereas we uniformly sample weight decays between $0$ and $10^{-5}$.

\fi

\end{document}